\documentclass[letterpaper, 10 pt, journal, twoside]{IEEEtran}

\title{LineRides: Line-Guided Reinforcement Learning for Bicycle Robot Stunts}

\author{
Seungeun Rho$^{1,2}$, 
    Shamel Fahmi$^{1}$, 
    Jeonghwan Kim$^{1,2}$, 
    Arianna Ilvonen$^{1}$
    Sehoon Ha$^{2}$, 
    and Gabriel Nelson$^{1}$%
    \thanks{Manuscript received: January, 15, 2026; Revised April, 1, 2026; Accepted April, 28, 2026.}
    \thanks{This paper was recommended for publication by Editor Lucia Pallottino upon evaluation of the Associate Editor and Reviewers' comments.}
    \thanks{The authors are with 
    $^{1}$RAI Institute, Cambridge, MA, USA, and
    $^{2}$Georgia Institute of Technology, Atlanta, GA, USA.}
    \thanks{
    Project page: \href{https://seungeunrho.github.io/LineRides}{\texttt{seungeunrho.github.io/LineRides}}}
    \thanks{Digital Object Identifier (DOI): see top of this page.}}

\usepackage{cite} 
\usepackage{graphicx} 
\graphicspath{{./includes/}, {./includes/supplementary_includes/}}
\usepackage[]{units} 
\usepackage{xspace} 
\usepackage{amsfonts} 
\usepackage[dvipsnames]{xcolor} 
\usepackage[final]{hyperref} 
\hypersetup{colorlinks=true, citecolor=MidnightBlue, linkcolor=MidnightBlue, urlcolor=MidnightBlue}
\usepackage{tabularx}
\usepackage{multicol}
\usepackage{multirow}
\usepackage{subcaption}  
\usepackage{float}
\usepackage[nonumberlist, nogroupskip, section=section, numberedsection=autolabel]{glossaries} 
\newacronym{rl}{RL}{Reinforcement Learning}
\newacronym{lineride}{LineRides}{LineRides}
\newacronym{umv}{UMV}{Ultra Mobility Vehicle}
\newcommand{\skill}[1]{{\fontfamily{qcr}\selectfont #1}}
\newcommand{\etal}{et~al.\xspace}

\begin{document}

\markboth{IEEE Robotics and Automation Letters. Preprint Version. Accepted May, 2026}
{Rho \MakeLowercase{\textit{et al.}}: LineRides: Line-Guided RL for Bicycle Robot
Stunts} 

\maketitle

\begin{abstract}
Designing reward functions for agile robotic maneuvers in reinforcement learning remains difficult, and demonstration-based approaches often require reference motions that are unavailable for novel platforms or extreme stunts. We present \acrshort{lineride}, a line-guided learning framework that enables a custom bicycle robot to acquire diverse, commandable stunt behaviors from a user-provided spatial guideline and sparse key-orientations, without demonstrations or explicit timing. \acrshort{lineride} handles physically infeasible guidelines using a tracking margin that permits controlled deviation, resolves temporal ambiguity by measuring progress via traveled distance along the guideline, and disambiguates motion details through position- and sequence-based key-orientations. We evaluate \acrshort{lineride} on the \gls{umv} and show that the policy trained with our methods supports seamless transitions between normal driving and stunt execution, enabling five distinct stunts on command: \skill{MiniHop}, \skill{LargeHop}, \skill{ThreePointTurn}, \skill{Backflip}, and \skill{DriftTurn}.
\end{abstract}

\section{Introduction}
Training reinforcement learning (RL) policies to perform agile maneuvers on robots is inherently challenging, as it is difficult to design reward functions that accurately capture the intended motion. Even for a seemingly simple skill such as \emph{jumping}, which is conceptually intuitive, translating this intuition into effective reward terms is nontrivial. 
Prior works have explored a wide range of reward formulations, including contact-based penalties, waypoint tracking, and phase-specific reward shaping~\cite{zhuang2023robot, li2023robust, cheng2024extreme}. However, these rewards are often task-specific and not easily generalizable to new behaviors. 

Alternative motion-imitation approaches leverage demonstrations to guide learning~\cite{peng2018deepmimic, RoboImitationPeng20, li2023learning, tessler2024maskedmimic, rho2025reference}.
These methods are effective when high-quality references are available. 
However, references for novel robots or extreme maneuvers and stunts may be difficult to acquire. 
For example, when demonstrations originate from different embodiments, such as other robots or animals~\cite{kang2025learning}, retargeting introduces additional sources of error, especially when the target robot has a unique structure such as~\gls{umv} shown in Fig.~\ref{fig:UMVMini}.

In this work, we introduce \acrshort{lineride}, a general learning framework that enables robots to acquire versatile stunt behaviors from a simple user-provided \emph{line} and a sequence of \emph{key-orientations}.

\begin{figure}\centering
\includegraphics[width=1.0\columnwidth]{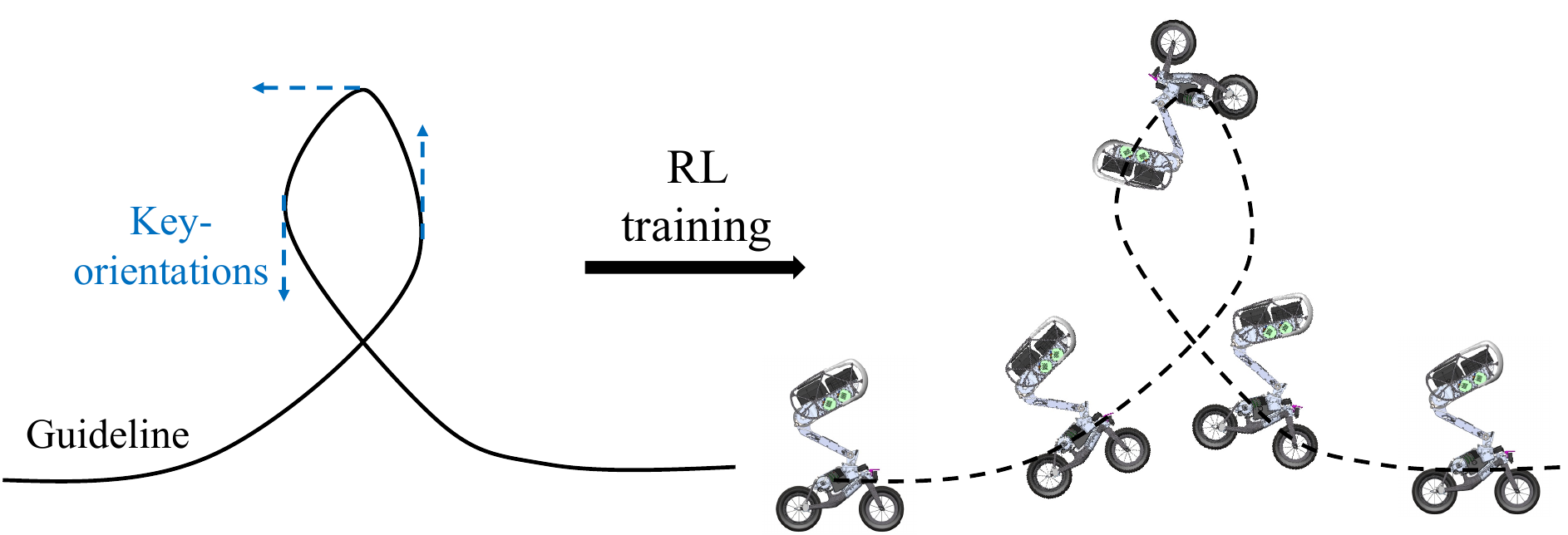}
\caption{Once the desired behavior is specified as a line and key-orientations, \acrshort{lineride} learns to accomplish the corresponding stunt maneuvers.}
\label{fig:guideline_intro}
\end{figure}

The \emph{line} provides an intuitive way to specify the desired motion. Users can define virtually any stunt trajectory using either a parameterized function, such as a Hermite curve, or a trajectory optimization based on simplified dynamics models. As illustrated in Fig.~\ref{fig:guideline_intro}, once a line is given, the robot learns to follow it, thereby realizing the intended agile motion. We refer to this user-specified line as the \emph{guideline}. 

Designing a method to follow such a guideline involves three key challenges.
First, the user-provided gudieline could be physically infeasible. For example, to draw a line for defining backflip stunt illustraed in Fig.~\ref{fig:guideline_intro}, user cannot know of the right height, width, and curvature of the curve. To deal with this, we introduce a \emph{margin parameter} that allows a controlled amount of deviation from the line, 
providing the policy with flexibility to satisfy physical constraints while still following the intended path. 

Second, unlike conventional demonstration-based approaches~\cite{peng2018deepmimic,peng2022ase,tessler2023calm}, our guideline does not include explicit temporal information. As a result, it is unclear \emph{a priori} when the robot should reach each point along the guideline. This introduces ambiguity in determining appropriate early termination conditions during training. Since early termination plays a critical role in training efficiency, this ambiguity poses a significant challenge. To address this issue, we use the robot’s cumulative traveled distance as a progression measure along the guideline, enabling consistent target selection without requiring explicit timing information. We describe this mechanism in detail in Section~\ref{section:methods}.

Finally, while the guideline specifies the overall spatial trajectory, it does not encode orientation information. Orientation is often critical for controlling fine-grained motion details, such as distinguishing between rear-wheel-first and front-wheel-first landings in the mini-hop example shown in Fig.~\ref{fig:guideline_tracking}. To address this limitation, we introduce \emph{key-orientations}, which specify desired base orientations at selected waypoints along the guideline. We consider two types of key-orientations. \emph{Position-based} key-orientations associate a target orientation with a specific waypoint, while \emph{sequence-based} key-orientations define a sequence of orientations that guide smooth transitions between two position-based key-orientations. Both formulations are described in detail in Section~\ref{sec:key-orientations}.

In addition, stunt skills are only meaningful when integrated with normal driving behavior. To this end, \acrshort{lineride} trains a single policy in an end-to-end manner that supports both \emph{driving mode} and \emph{stunt mode}. In driving mode, the robot maintains ground contact and follows user-issued joystick commands for basic locomotion, including forward and backward motion, turning, acceleration, and deceleration. Upon pressing a designated button, the robot transitions into stunt mode, executes the stunt behavior defined by the guideline, and seamlessly returns to driving mode upon completion.

We validate our framework across a diverse set of stunt skills. In simulation, we train challenging maneuvers such as drift turns and backflips, while on real hardware, we demonstrate three-point turns and two levels of jumping. The results show that \acrshort{lineride} offers a generic framework for acquiring agile behaviors on~\gls{umv}.

In summary, our contributions are as follows:
\begin{itemize}
    \item We present \acrshort{lineride}, a general reinforcement learning framework that enables robots to acquire diverse stunt behaviors from a simple user-provided spatial guideline, without requiring joint-level demonstrations.
    \item We propose a traveled-distance-based termination criterion that enables early termination without requiring time-coupled data.
    \item We introduce key-orientations as an explicit mechanism for resolving orientation ambiguity and controlling fine-grained motion details along the guideline.
    \item We demonstrate the effectiveness of \acrshort{lineride} across five challenging stunt behaviors in simulation and three real-world behaviors on the \gls{umv} platform.
\end{itemize}

\section{Related Work}

\subsection{Learning Agile Stunt Maneuvers}

With the advancement of reinforcement learning (RL), both legged and wheeled robots have demonstrated the ability to learn highly agile stunt maneuvers. 
Legged robots have shown capabilities in learning dynamic and acrobatic behaviors~\cite{rhounsupervised, zhuang2023robot, cheng2024extreme, hoeller2024anymal}, 
while wheeled platforms have also learned diverse and agile motion skills~\cite{zheng2022continuous, baltes2023deep, wang2024bayesian, bjelonic2022survey, kim2026flip}. 
Beyond skill learning, legged robots have achieved robustness across various terrains~\cite{he2025attention, kim2025high} and even exhibited high-speed running near their physical limits~\cite{miller2025high, margolis2024rapid}. 
More recently, bipedal robots have also demonstrated the ability to acquire a wide range of agile behaviors~\cite{xie2025kungfubot, he2025omnih2o, chen2025gmt, he2025asap, li2025reinforcement}.

Here, we highlight how diverse efforts have been required to enable robots to learn agile behaviors. 
For instance, to train a relatively simple \emph{jump} skill, \cite{zhuang2023robot} introduced a virtual obstacle and penalized the robot’s body overlap with it to induce a jumping motion. 
\cite{li2023robust} designed a phase-based reward that distinguishes between pre-landing and post-landing stages, assigning distinct objectives to each phase. 
\cite{cheng2024extreme} shaped the behavior using human-specified waypoints as intermediate guidance signals. 
To reduce the reliance on such hand-crafted rewards, \cite{rhounsupervised} adopted unsupervised skill discovery for learning diverse locomotion behaviors, although the agility and expressiveness of the resulting skills remained limited.

Overall, these studies explore a wide spectrum of reward design and training paradigms for agile locomotion. 
In contrast, our work introduces a general learning framework that can learn different stunt skills through a single motion-representation based on \emph{lines}.

\subsection{Lines as Goal Describing Modality}

\subsubsection{Manipulation}
\emph{Lines} or \emph{waypoints} have emerged as an expressive modality for specifying goals in recent robotic manipulation. Gu et al.~\cite{gu2023rt} first proposed a robot-arm control policy conditioned on trajectory sketches. 
Similarly, \cite{zhi2024instructing} and \cite{yu2025sketch} leveraged human-drawn sketches as an intuitive representation of task intent, providing dense goal information. 
\cite{mehta2024waypoint, mehta2025l2d2} further introduced waypoint-based RL frameworks, where agents learn high-level waypoint sequences instead of low-level motor commands. 
Inspired by these advances, we aim to extend this idea to the domain of agile locomotion and stunt skill learning.


\subsubsection{Locomotion}
Several recent works have explored geometric interfaces for specifying locomotion behaviors. 
Bussola~\etal use Bézier curves to guide quadruped jumping~\cite{bussola2025guided}, but in their framework the curve serves as a high-level action that is converted into joint trajectories via inverse kinematics and then tracked by a PD controller, rather than as a target behavior specification itself.

WASABI~\cite{li2023learning} and RobotKeyFraming (RKF)~\cite{zargarbashi2024} are more closely related to our setting. 
WASABI learns from partial demonstrations represented as time-indexed observation sequences, while RKF specifies motions as time-assigned keyframes. 
In both cases, the user must specify not only the desired motion, but also its temporal schedule. 
This can be difficult for highly dynamic stunt behaviors, where the required timing and velocity profile are not known in advance and may vary across initial conditions. 
In contrast, \acrshort{lineride} is purely position-based: the user specifies only a spatial guideline, and the reinforcement learning process discovers the timing and control strategy needed to realize the stunt.

\section{The Ultra Mobility Vehicle~(UMV)} 

\begin{figure}\centering
\includegraphics[width=0.7\columnwidth]{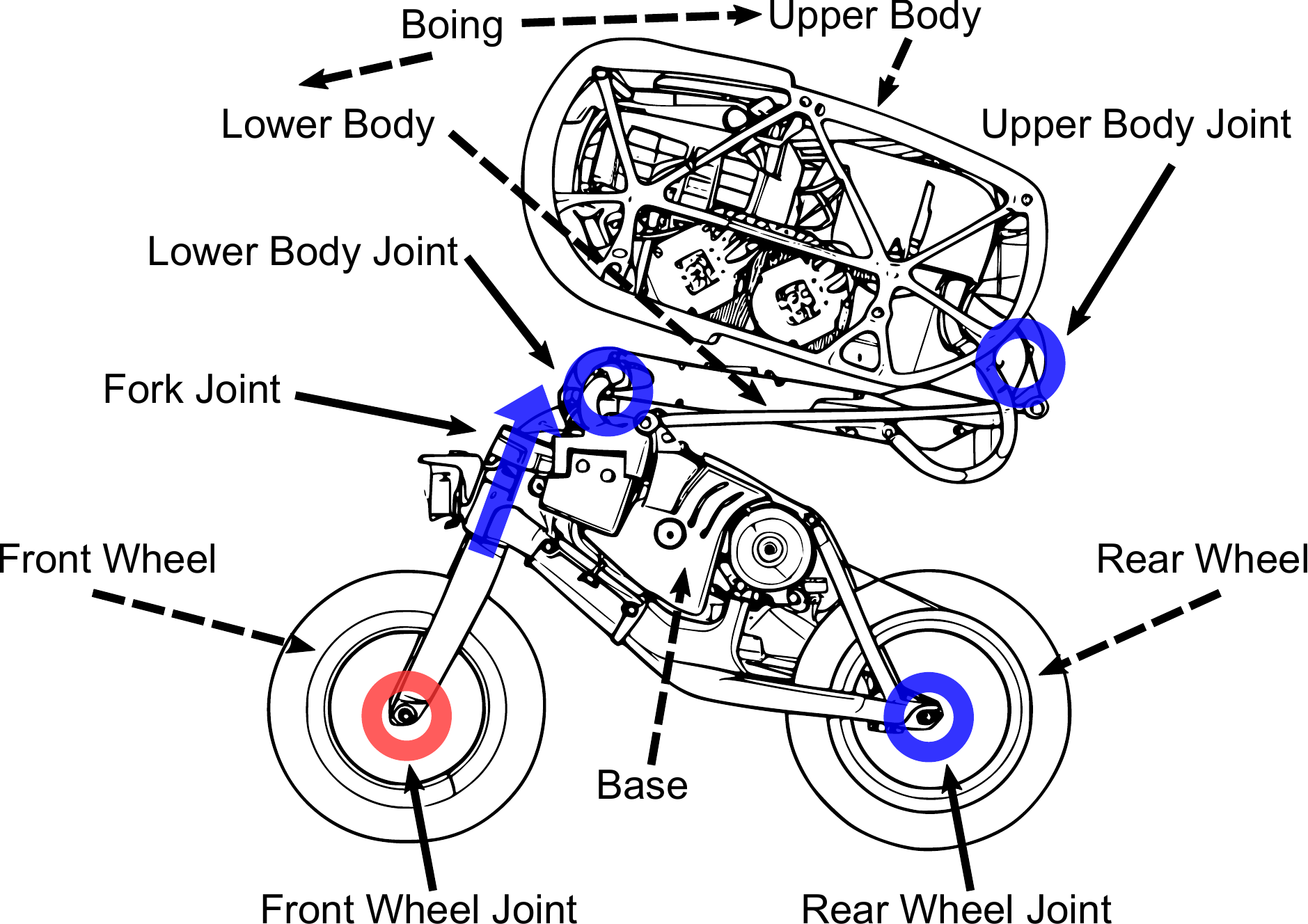}
\caption{A 2D overview of the \gls{umv} model used in this work.}
\label{fig:UMVMini}
\end{figure}

\gls{umv} is a custom-built, two-wheeled robot that resembles a child-sized bicycle (Fig.~\ref{fig:UMVMini}). 
Unlike a conventional bike, it features a large articulated module, referred to as \emph{boing}, mounted on top of the frame. 
Boing can be actuated vertically—either lifted upward or contracted downward—through two joints, namely the \emph{upper-body} and \emph{lower-body} joints. 
This module contains the majority of the robot’s mass (approximately 18\,kg out of the total 26\,kg), including the batteries and actuators. 
By shifting this concentrated mass, the robot can generate substantial momentum, enabling highly dynamic and acrobatic maneuvers. 

The remainder of the robot follows a conventional bicycle design: the fork joint steers the front wheel, the rear wheel provides driving torque, and the front wheel remains passive. In total, the version of the \gls{umv} used in this work has four actuated joints. Overall, the addition of the \emph{boing} transforms an otherwise simple bicycle frame into a platform capable of agile, stunt-like behaviors.

\section{Problem Formulation}

We formulate the stunt skill learning problem for \gls{umv} as a Markov Decision Process (MDP), defined as
\[
\mathcal{M} = \{\mathcal{S}, \mathcal{A}, \mathcal{R}, \mathcal{P}, \gamma\},
\]
where $\mathcal{S}$ denotes the state space, $\mathcal{A}$ the action space, $\mathcal{R}$ the reward function, and $\mathcal{P} = \Pr(s' \mid s, a)$ the transition dynamics that describe the probability of reaching the next state $s'$ given the current state $s$ and action $a$. The scalar $\gamma \in [0, 1]$ is the discount factor.

The agent follows a stochastic policy $\pi(a \mid s)$, parameterized by a neural network with parameters $\theta$, denoted as $\pi_\theta$. The goal of reinforcement learning is to optimize $\theta$ to maximize the expected discounted return:
\begin{equation}
\mathcal{J}(\theta) = \mathbb{E}_{\pi, \mathcal{P}} \left[ \sum_{t=0}^{\infty} \gamma^t r_t \right].
\end{equation}
We adopt Proximal Policy Optimization (PPO)~\cite{schulman2017proximal} as the RL algorithm for training $\pi_\theta$.

\section{\acrshort{lineride}}
\label{section:methods}

In this section, we describe how the provided line and key-orientations are used to enable the robot to perform agile stunt behaviors. We then explain how the guideline can be generated, and finally discuss how stunt execution is integrated into the normal driving mode and activated by a human trigger.

\subsection{Tracking Guideline}
\label{sec:guideline_tracking}

Our objective is to enable the robot to execute a stunt behavior represented by a guideline
\( l = [p_1, p_2, \ldots, p_n] \),
expressed in the robot’s local coordinate frame, where each waypoint \( p_i \in \mathbb{R}^3 \).

\textbf{Rewards.}
At each timestep, a single waypoint \( p_i \) is selected as the active target \( p_{\text{goal}} \), and rewards are computed with respect to this target. When the robot's position \( x^{\text{stunt}}_{\text{t}} \) reaches the current target \( p_{\text{goal}} \), the active target is updated to the next waypoint \( p_{i+1} \).
Given \( p_{\text{goal}} \), we define the tracking reward as the negative change in distance to the target:
\begin{equation}
r_t^{\text{line}} = - \left(\|x^{\text{stunt}}_{\text{t}} - p_{\text{goal}}\|_2 - \|x^{\text{stunt}}_{t-1} - p_{\text{goal}}\|_2\right),
\end{equation}
which yields a positive reward whenever the robot moves closer to the target waypoint.

\begin{figure}\centering
\includegraphics[width=0.99\columnwidth]{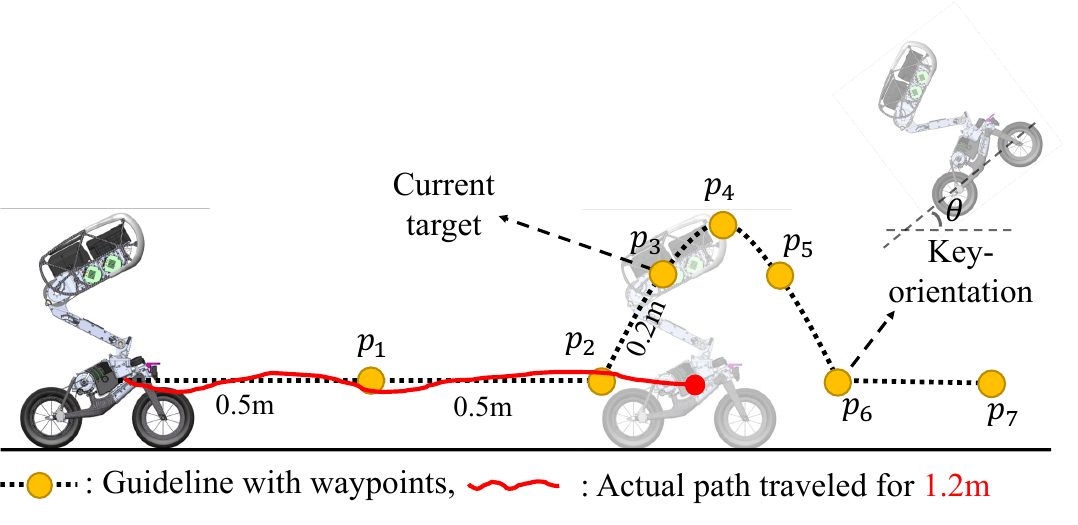}
\caption{An example guideline for a mini-hop motion with 7 waypoints and one key-orientation. The target waypoint is determined based on the traveled distance along the path.}
\label{fig:guideline_tracking}
\end{figure}

\textbf{Early Termination.}
Since the guideline provides no explicit temporal information, determining when to terminate an episode is nontrivial. Instead, we leverage the known cumulative distance required to reach each waypoint and infer progression along the guideline based on the \emph{distance traveled} by the robot.

For each waypoint \( p_i \), we precompute the cumulative arc-length of the guideline up to that waypoint, denoted by \( d_i \). The guideline can thus be redefined as
$$
l = [(p_1, d_1), (p_2, d_2), \ldots, (p_n, d_n)],
$$
where \( d_1 < d_2 < \cdots < d_n \).

During training, we continuously integrate the distance traveled by the robot’s base, denoted by \( d_{\text{t}} \). When the \( i \)-th waypoint \( p_i \) is the active target and the condition \( d_{\text{t}} > d_i \) is satisfied, we evaluate whether the robot has successfully reached \( p_i \). If the waypoint has not been reached, the episode is terminated. As illustrated in Fig.~\ref{fig:guideline_tracking}, if the robot has traveled \(1.2\,\mathrm{m}\) and \( d_3 = 1.2\,\mathrm{m} \), we terminate the episode if robot deviate too far from $p_3$.

\textbf{Margin.}
Because user-provided guidelines may not correspond to physically feasible trajectories, we introduce a margin parameter \( d_{\text{margin}} \) that provides tolerance for deviations from the guideline.

In detail, a waypoint is considered reached when the distance between the robot’s current position \( x^{\text{stunt}}_{\text{t}} \) and the target \( p_{\text{goal}} \) is below a threshold \( d_{\text{margin}} \):
\begin{equation}
\|x^{\text{stunt}}_{\text{t}} - p_{\text{goal}}\|_2 < d_{\text{margin}}.
\end{equation}
A larger margin allows the policy to tolerate tracking error without premature termination, enabling robust learning even when exact adherence to the guideline is infeasible.

\subsection{Controlling Motion Details via Key-Orientation Tracking}
\label{sec:key-orientations}

Key-orientations specify the desired orientation of the robot’s base at selected waypoints along the guideline. As illustrated in Fig.~\ref{fig:guideline_tracking}, the user can encourage a rear-wheel-first landing, which helps absorb impact more effectively, by assigning a positive pitch angle at the waypoint at landing. Formally, key-orientations can be defined in two ways: \emph{position-based} and \emph{sequence-based}. 

\textbf{Position-Based.} The most straightforward approach is to associate each desired orientation directly with a spatial position. Each key-orientation \(k_i = (p_i, q_i)\) is defined as a tuple consisting of a waypoint \(p_i \in \mathbb{R}^3\) and an orientation \(q_i \in \mathbb{R}^4\) represented as a quaternion. 

When the robot reaches sufficiently close to \(p_i\), we compute the angular difference $\theta^{\text{diff}}$ between its current orientation \(q_t\) and the target orientation \(q_i\), and assign a reward that increases as this difference decreases:
\begin{equation}
r_t^{\text{rot}} = \exp(-\theta^{\text{diff}}) = \exp(2\,\arccos\!\left( \left| q_t \cdot q_i \right| \right)).
\end{equation}

Additionally, we terminate the episode when the robot’s orientation deviates excessively from the target. Specifically, if
$
\theta^{\text{diff}} > \theta^{\text{thres}},
$
the episode is terminated immediately.

\textbf{Sequence-Based.}
A limitation of the position-based approach is that it requires specifying the exact positions of desired orientations \emph{before training}, which is often difficult or impossible. For example, in a backflip, we may want the robot to pass through \(90^\circ\), inverted, and \(270^\circ\) poses before returning upright, but the exact positions of these poses depend on the robot’s dynamics and are hard to know in advance.

To overcome this limitation, we introduce \emph{sequence-based} key-orientations, which allow omitting explicit spatial positions for intermediate key-orientations between the position-based ones, as follows:
\begin{equation}
k = \{(p_\text{init}, q_\text{init}),\, q_1, q_2, \ldots, q_n,\, (p_\text{fin}, q_\text{fin})\}.
\end{equation}

These sequence-based key-orientations are not necessarily tied to explicit spatial coordinates but instead define a temporal sequence of desired orientations. 
This formulation assumes that the angular gap between the robot’s base orientation and the current target orientation \(q_i\) should decrease monotonically over time, as in the backflip example.

The reward is computed based on the angular progress toward the current target orientation. 
Let $\theta^{\text{diff}}_t$ denote the angle between the robot’s current orientation \(q_t\) and the target orientation \(q_i\). 
The reward is then defined as
$$
r_t^{\text{rot}} = -(\theta^{\text{diff}}_t - \theta^{\text{diff}}_{t-1}),
\ \ \text{where} \ \ 
\theta^{\text{diff}}_t = 2\,\arccos\!\left(\left| q_t \cdot q_{i} \right|\right),
$$
which yields a positive reward when the robot’s orientation moves closer to the target key-orientation. 
Early termination is applied when monotonicity is violated; if the orientation difference increases, the episode terminates.

\subsection{Drawing Guidelines}

\subsubsection{Hermite Curve}

To generate smooth and diverse spatial lines using user-specified parameters, we employ a cubic Hermite curve representation. 
Hermite curves allow explicit control over both position and tangent (velocity) at each endpoint, enabling intuitive shaping of the desired trajectory.

Given two points \(x_0, x_1 \in \mathbb{R}^3\) and their corresponding tangent vectors \(m_0, m_1 \in \mathbb{R}^3\), the Hermite curve \(p(u)\) for \(u \in [0, 1]\) is defined as
\begin{multline}
p(u) = (2u^3 - 3u^2 + 1)\,x_0 
      + (u^3 - 2u^2 + u)\,m_0 \\[2pt]
      + (-2u^3 + 3u^2)\,x_1 
      + (u^3 - u^2)\,m_1 .
\end{multline}

Because the curve is parameterized, we can sample as many intermediate waypoints as needed. 
In practice, we first densely sample approximately 1{,}000 points to compute the empirical cumulative distance $d_i$ for each \(p_i \in l\). 
We then re-sample a smaller subset (typically 10–20 waypoints) to form the final guideline \(l\).

\subsubsection{Trajectory optimization with simplified model}

\begin{figure}\centering
\includegraphics[width=0.75\columnwidth]{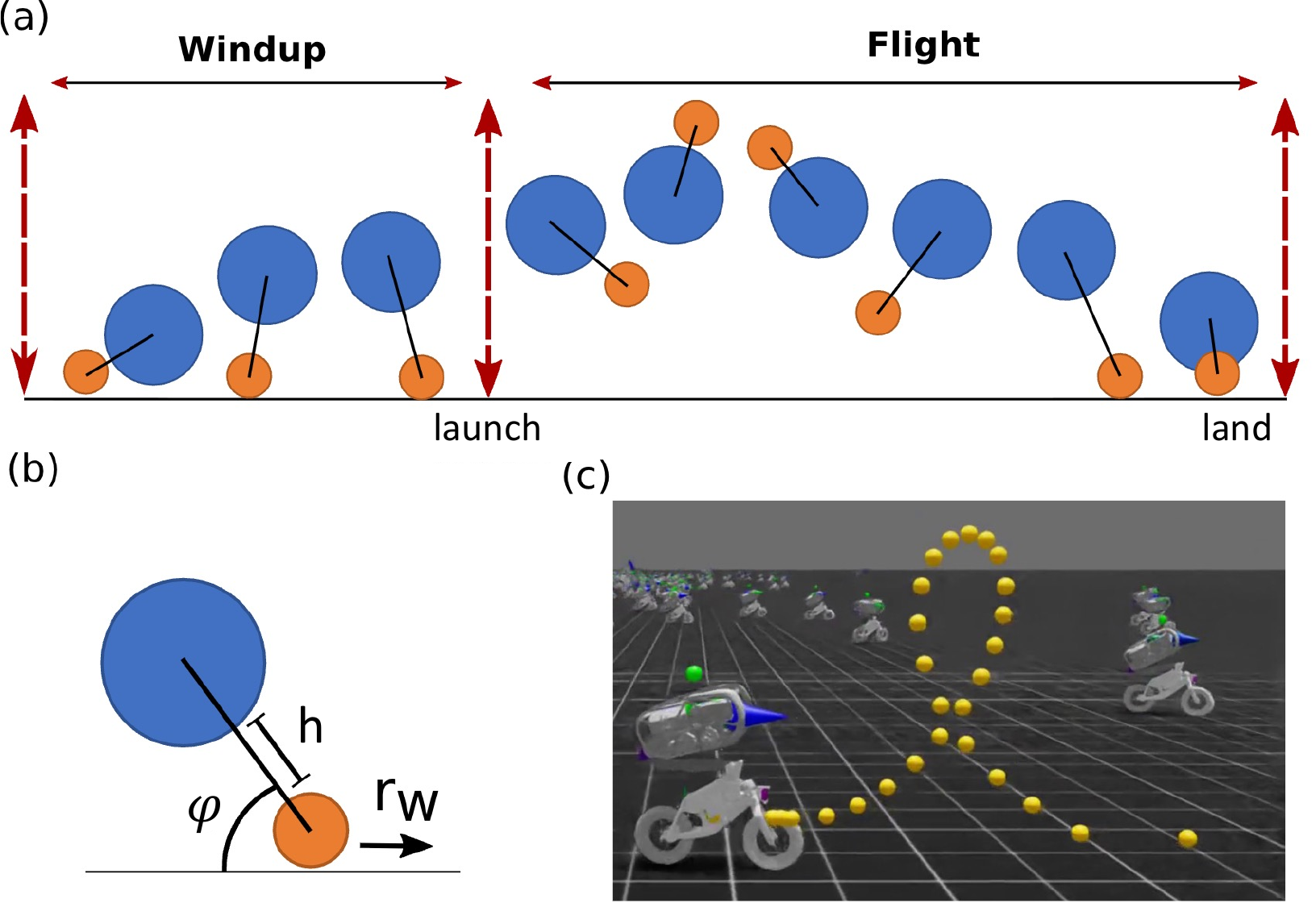}
\caption{Trajectory optimization with simplified two-mass models can provide guidelines. (a) Simulated backflip using a two-mass model. (b) Two-mass model, including degrees of freedom. (c) Backflip trajectory from the full simulated robot.}
\label{fig:two_mass_line}
\end{figure}

We can also generate a guideline through trajectory optimization with a simplified dynamic model. 
For the backflip, we model the bike-like robot as a two-mass system connected by a prismatic joint, as shown in Fig.~\ref{fig:two_mass_line}. The state and control are
\begin{multline}
x = [x_{\text{com}}, z_{\text{com}}, \dot x_{\text{com}}, \dot z_{\text{com}}, \phi, \dot\phi, h, \dot h] \in \mathbb{R}^{8}, \\
u = [\ddot{h}, \tau, r_w] \in \mathbb{R}^{3}.
\end{multline}
We solve a two-phase optimization problem with ground-contact and flight phases, and use the resulting base trajectory directly as the guideline for backflip training.

\subsection{Combining Stunt Execution with Driving Mode}

While the proposed method enables the robot to execute agile stunt behaviors, it does not by itself determine when these behaviors should be activated. In realistic operation, the robot should remain in a stable driving mode by default and initiate a stunt only when explicitly commanded by the user, with smooth transitions between modes. To support this, \acrshort{lineride} trains a single end-to-end policy that takes desirable mode, either \emph{driving} and \emph{stunt} as user input. 
This user input is randomly instantiated during training, and each mode comes with designated reward function and termination conditions.

\subsubsection{Training in Driving Mode}

In driving mode, the robot tracks three types of user commands: a target forward linear velocity $v^{\text{drive}}$, a target yaw angular velocity $\omega^{\text{drive}}$, and desired upper- and lower-body joint angles $\rho^{\text{drive}}$.
Each command is sampled from a predefined distribution with its own resampling interval, exposing the robot to a diverse combination of command inputs.

The reward function is designed to minimize the deviation between the robot’s actual motion and commanded targets. Regularization penalties are added to improve robustness for both of the modes. The complete reward formulation, command ranges, and command sampling intervals are summarized in Table~\ref{tab:driving_mode_rew}. 

The linear velocity command can take both zero and negative values. A negative value instructs the robot to drive backward, while a zero value corresponds to a \emph{track-stand} behavior, where the robot maintains balance without forward motion through active fork control.

\begin{table}[t]
\caption{Rewards for Both Stunt and Driving Modes}
\label{tab:driving_mode_rew}
\centering
\resizebox{\columnwidth}{!}{%
\begin{tabular}{lccc}
\hline
\textbf{Component} & \textbf{Formulation}  & \textbf{Range} & \textbf{Interval(s)} \\

\hline
\multicolumn{4}{c}{\textit{Stunt Mode Rewards}} \\
Guideline Track & $\|x^{\text{stunt}}_{\text{t}} - p_{\text{goal}}\|_2 - \|x^{\text{stunt}}_{t-1} - p_{\text{goal}}\|_2$ & -- & -- \\
\multirow{2}{*}{Key-orientation Track} &
$\exp\!\left(2\,\arccos\!\left( \left| q_t \cdot q_i \right| \right)\right)$
& -- & -- \\
& or, $-(\theta^{\text{diff}}_t - \theta^{\text{diff}}_{t-1})$ & & \\  

\hline
\multicolumn{4}{c}{\textit{Driving Mode Rewards}} \\
Lin-vel Track & $3*\exp(\| v^{\text{base}} - v^{\text{drive}} \|^2)$ & 
$\operatorname{U}[-1, 2]$ & $\operatorname{U}[2, 15]$
\\

Ang-vel Track & $3*\exp(\| \omega^{\text{base}} - \omega^{\text{drive}} \|^2)$ & 
$\operatorname{U}[-1.5, 1.5]$ & $\operatorname{U}[3, 8]$\\

Boing-pos Track & $3*\exp(\| \rho^{\text{boing}} - \rho^{\text{drive}} \|^2)$ & $ \operatorname{U}[0.06, 1.0]$ & $\operatorname{U}[3, 10]$ \\
[2pt]\hline

\multicolumn{4}{c}{\textit{Regularization Penalties for Both Modes}} \\
Action Smoothness & $-10^{-4}*\| a_t - a_{t-1} \|^2$ & -- & -- \\

Fork Velocity & $-0.001*\dot{q}_{\text{fork}}^2$ & -- & -- \\

Contact Force & $-10^{-6}*\| max(F_{\text{contact}}-350, 0) \|^2$ & -- & -- \\

Body Joint Limits & $-\mathbb{I}(q_{\text{joint}} \notin [q_{\text{min}}, q_{\text{max}}])$ & -- & -- \\


Velocity Penalty & $-3*\| max(v_{\text{base}}-2, 0) \|^2$ & -- & -- \\
[2pt]\hline
\end{tabular}
}%
\vspace*{-2mm}
\end{table}

\subsubsection{Mode Transitions}

During training, all robot begins in driving mode, and after a random duration $t_{\text{switch}} \sim \operatorname{Uniform}(2, 5)$ seconds, the agent switches to stunt mode. At this point, the reward function and termination conditions are replaced with those defined for the stunt mode as well. The robot is informed of the current mode through a mode variable $c \in o_t$. During deployment, this value is given by the user through a joystick button trigger.

Once the agent enters stunt mode, it returns to driving mode only after the robot successfully reaches the final waypoint of the guideline. After completion, a new $t_{\text{switch}}$ value is sampled, and after the duration elapses, the robot re-enters stunt mode. Each episode lasts 20 seconds, during which the robot typically performs two to five stunt executions depending on the sampled switch duration.

\subsection{Observations, Actions, Control, and Domain Randomization}

\textbf{Observations:} The robot’s observation at time step $t$ is defined as
$$
o_t = [q, \dot{q}, \omega, g, a_{t-1}, c, v^{\text{drive}}, \omega^{\text{drive}}, \rho^{\text{drive}}, x^{\text{stunt}}_{\text{t}}] \in \mathbb{R}^{25}
$$
where $q \in \mathbb{R}^4$ denotes the actuated joint positions, 
$\dot{q} \in \mathbb{R}^4$ the joint velocities, 
$\omega \in \mathbb{R}^3$ the base angular velocity, 
$g \in \mathbb{R}^3$ the gravity vector expressed in the robot’s base frame, 
$a_{t-1} \in \mathbb{R}^4$ the previous action, and $c \in \{0, 1\}$ the user command indicating whether the robot is in \emph{driving mode} or \emph{stunt mode}. 

In the driving mode, the user command for driving such as 
$v^{\text{drive}} \in \mathbb{R}^1$, $\omega^{\text{drive}}$ and $\rho^{\text{drive}} \in \mathbb{R}^2$ are active. These driving command values are masked to zero during stunt mode. Conversely, $x^{\text{stunt}}_{\text{t}} \in \mathbb{R}^3$ is active only in stunt mode and measures the robot’s base position relative to its position at the moment when stunt mode is triggered.

Note that the guideline and key-orientations are used only during training to compute the reward.
They are not provided as policy inputs during either training or testing.

\begin{figure*}[t]  
    \centering
    \includegraphics[width=0.9\textwidth]{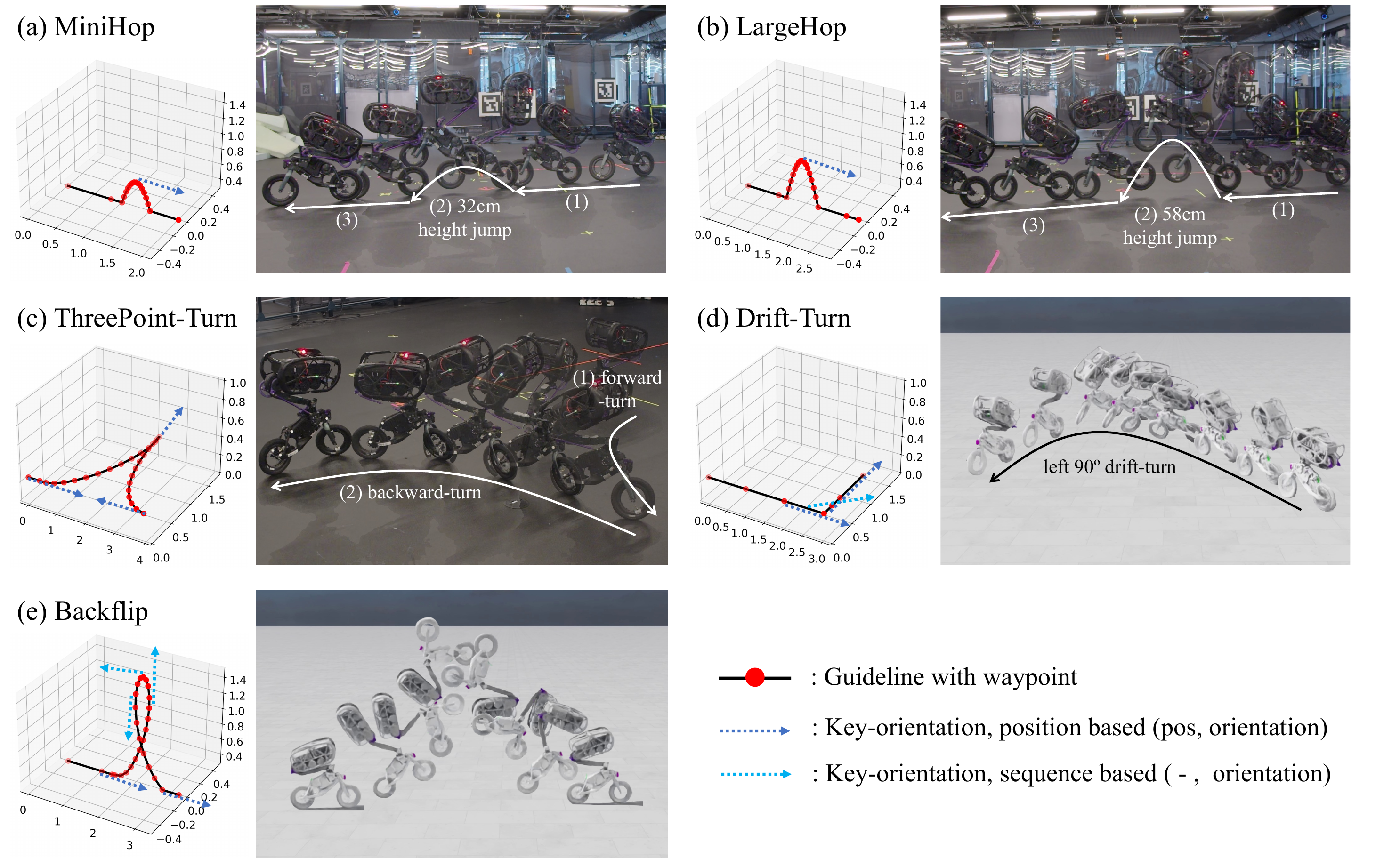}
    \caption{\acrshort{lineride} converts a user-drawn guideline (left) into a corresponding robot stunt maneuver (right). Cases (a), (b), and (c) are generated using cubic Hermite curves, (d) is drawn manually as it consists of simple straight lines, and (e) is produced via trajectory optimization. All five stunt skills are successfully trained and demonstrated. }
    \label{fig:learned_skills}
    \vspace*{-2mm}
\end{figure*}

\textbf{Actions:} The policy outputs a four-dimensional continuous action vector
$$
a_t = [a_{\text{upper-body}}, a_{\text{lower-body}}, a_{\text{fork}}, a_{\text{rearwheel}}] \in \mathbb{R}^4.
$$
It specifies desired joint position setpoints for the upper-body, lower-body, and fork joints, which have well-defined joint limits, and a velocity target for the continuously rotating rear wheel, whose absolute joint angle is unbounded.

\textbf{Control:} The final joint torques $\tau$ are computed using a proportional–derivative (PD) controller:
\begin{equation}
\tau = k_p (q_{\text{des}} - q) + k_d (\dot{q}_{\text{des}} - \dot{q}),
\end{equation}
where $q_{\text{des}}$ and $\dot{q}_{\text{des}}$ denote the desired joint position and velocity setpoints, respectively.

\begin{table}[t]
\caption{Domain Randomization Parameters}
\label{tab:domain_rand}
\centering
\begin{tabular}{lcc}
\hline

\textbf{Parameter} & \textbf{Range / Distribution} & \textbf{Unit} \\
\hline
Added Mass & $\operatorname{U}$[0.9, 1.1] & ratio \\
Friction Coefficient & $\operatorname{U}$[0.7, 1.5] & - \\
Motor Strength & $\operatorname{U}$[0.85, 1.05] & ratio \\
Actuator Gains Scale & $\operatorname{U}$[0.85, 1.15] & ratio \\
Actuation Delay & $\operatorname{U}$[0, 1] & steps \\
Obs. Noise (Joint Pos) & $\mathcal{N}(0, 0.001)$ & rad \\
Obs. Noise (Joint Vel) & $\mathcal{N}(0, 0.1)$ & rad/s \\
Obs. Noise (Ang Vel) & $\operatorname{U}$[-0.1, 0.1] & rad/s \\
Obs. Noise (Proj Gravity) & $\operatorname{U}$[-0.015, 0.015] & - \\
Body Velocity Disturbance & $\operatorname{U}$[-0.1, 0.1]  & m/s \\
\hline
\end{tabular}
\vspace*{-2mm}
\end{table}

\textbf{Domain Randomization:}
To reduce the sim-to-real gap, we apply domain randomization~\cite{tobin2017domain, peng2018sim} in both driving and stunt modes. The full set of randomized parameters is given in Table~\ref{tab:domain_rand}.



\section{Experimental Results}
\label{sec:experiments}

We train our policy in the high-throughput, GPU-based simulator IsaacLab~\cite{mittal2025isaac}, and then deploy it on real hardware for evaluation. Depending on the task, training takes 12--24 hours on an NVIDIA L40 GPU.

\subsection{\acrshort{lineride} Can Learn Diverse Stunt Maneuvers}
\label{sec:exp_stunts}
To evaluate the capability of \acrshort{lineride} in learning diverse stunt skills, 
we trained and tested a separate control policy for each of five distinct bike maneuvers in both real-world and simulated environments:
\begin{itemize}
    \item \skill{MiniHop}: a small vertical hop reaching approximately 32\,cm in height and 50\,cm in distance (real, sim)
    \item \skill{LargeHop}: a larger vertical hop reaching about 56\,cm in height and 80\,cm in distance (real, sim)
    \item \skill{ThreePointsTurn}: a planar two-step pivot turn that requires precise balance adjustments (real, sim)
    \item \skill{DriftTurn}: a planar left drift-turn maneuver of roughly 90° involving controlled rear-wheel slip (sim)
    \item \skill{Backflip}: a full vertical backward flip executed while moving forward (sim)
\end{itemize}
For safety reasons, \skill{DriftTurn} and \skill{Backflip} were evaluated only in simulation, since failed hardware trials could significantly damage the robot.

The user-provided guidelines, along with the corresponding key-orientations and the resulting learned stunt behaviors, are visualized in Fig.~\ref{fig:learned_skills}. 
As shown in the figure, between one and five key-orientations were sufficient to specify each stunt motion. 

The learned skills exhibited strong robustness under repeated execution. In each driving trial, the robot was commanded to execute the consecutive stunt three to five times, consistently achieving a 100\% success rate. 
Please refer to the supplementary video for additional details. 

We also evaluated the robustness of the learned policy in driving mode. 
After completing each stunt, the robot seamlessly returned to driving mode and continued following the user’s driving commands. 
The driving tasks included a variety of maneuvers such as turns with varying angles, acceleration and deceleration phases, track-stand balancing (zero-velocity command), and backward driving. 
We confirmed that the driving policy remained highly stable and responsive, faithfully following user commands. 
A representative 50-second driving sequence is provided in the supplementary video.

\subsection{\acrshort{lineride} Can Handle Physically Infeasible Guidelines}

As discussed in Section~\ref{sec:guideline_tracking}, we introduce a \emph{margin} to account for potential infeasibility in user-drawn guidelines. A fixed margin of 30,cm is used across all stunt motions. To examine the effect of this design, we study how the robot behaves when a provided guideline is physically infeasible but still close to a feasible trajectory. Specifically, we record the realized base trajectory on hardware for five trials of \skill{MiniHop} and three trials of \skill{LargeHop}, and overlay the resulting trajectories with their corresponding guidelines.

\begin{figure}\centering
\includegraphics[width=0.9\columnwidth]{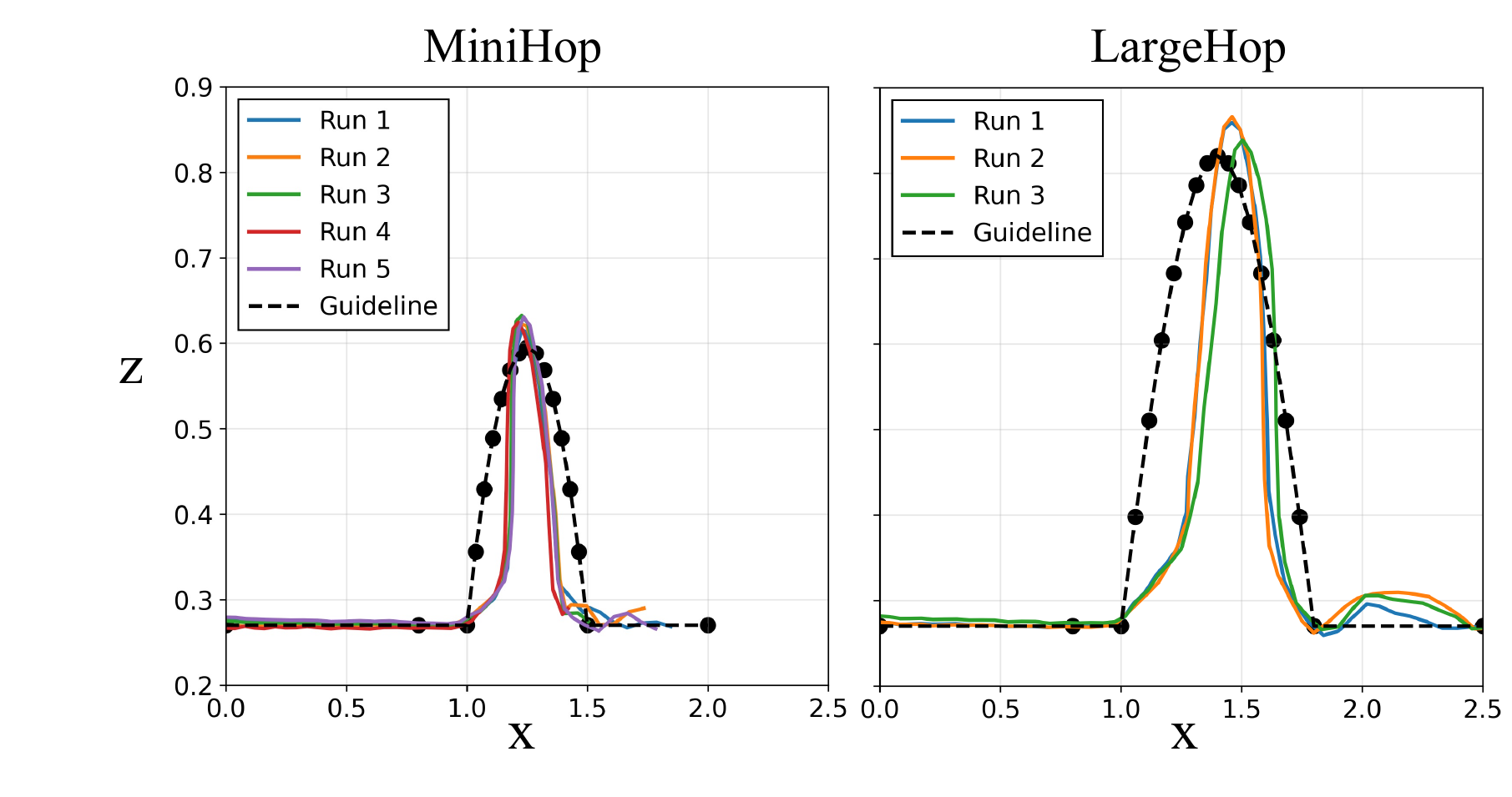}
\caption{Trajectory of the robot's base during hardware deployment, along with its guideline.}
\label{fig:hop_traj_in_real}
\vspace*{-3mm}
\end{figure}

As shown in Fig.~\ref{fig:hop_traj_in_real}, the realized trajectories deviate from the idealized guidelines. Nevertheless, with the help of the margin, the robot is able to successfully execute the intended maneuvers. This deviation illustrates that the tolerance margin allows the robot to depart from the guideline when necessary while still preserving the intended behavior. Moreover, the realized trajectories are highly consistent across repeated hardware trials, resulting in nearly overlapping curves, which indicates the robustness of the learned controller.

\subsection{\acrshort{lineride} Can Control Fine-Grained Motion Details}

Augmenting the guideline with key-orientations provides a mechanism for controlling subtle yet critical aspects of motion. For instance, to achieve a rear-wheel-first landing, the robot must coordinate its body extension and contraction to absorb impact efficiently. This can be encouraged by adding an additional key-orientation on the guideline right before landing. 

To evaluate this effect, we conducted an experiment comparing two policies trained with and without key-orientation supervision during the same \skill{MiniHop} maneuver. We measured the pitch angle of the robot’s base at a height of 15\,cm—corresponding to the moment when the rear wheel begins to make ground contact. Table~\ref{tab:effect_of_keypos} summarizes the commanded and realized pitch angles across three trials.

Without key-orientations, the robot’s pitch angle before landing varied between $-6^\circ$ and $-4.6^\circ$, indicating undesired front-wheel-first postures. In contrast, when a key-orientation explicitly specifying a $17^\circ$ pitch was provided, the resulting motions closely matched the intended orientation, with realized angles between $11^\circ$ and $13^\circ$. These results confirm that key-orientations enable precise modulation of motion semantics while maintaining the overall trajectory learned from the guideline.

\subsection{Baseline comparison}

\begin{figure}[t]
    \centering
    \includegraphics[width=0.95\columnwidth]{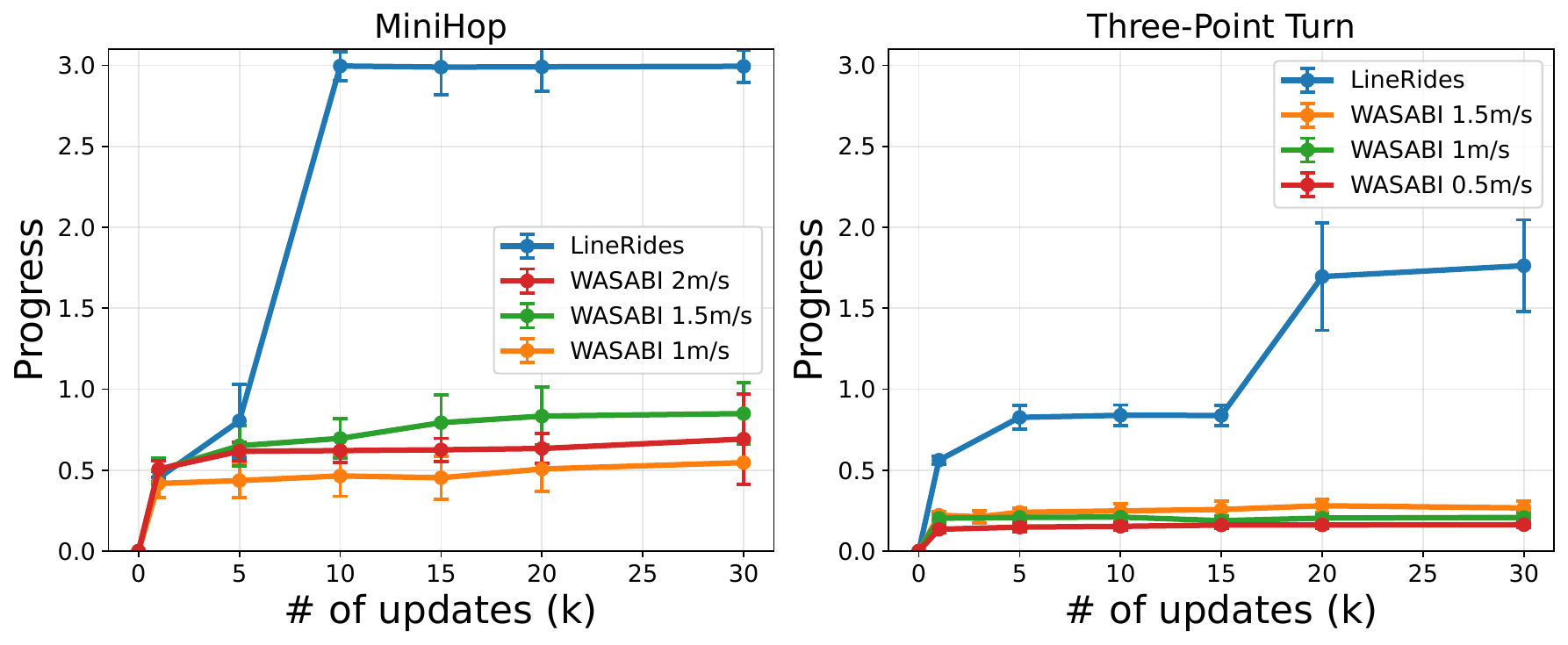}
    \caption{Comparison between \acrshort{lineride} and the WASABI baseline. Each policy attempted three consecutive stunts per episode; a score of 3 indicates success on all three.}
    \label{fig:baseline_wasabi}
\end{figure}

\begin{table}[t]
\centering
\caption{Effect of key-orientations on the Robot’s Pitch Angle}
\label{tab:effect_of_keypos}
\renewcommand{\arraystretch}{1.1}
\resizebox{\columnwidth}{!}{%
\begin{tabular}{l|ccc|ccc}
\hline
 & \multicolumn{3}{c|}{Without key-orientations} & \multicolumn{3}{c}{With key-orientations} \\
 & Run 1 & Run 2 & Run 3 & Run 1 & Run 2 & Run 3 \\
\hline
Target Orientation & -- & -- & -- & $17^\circ$ & $17^\circ$ & $17^\circ$ \\
Realized & $-5.7^\circ$ & $-6.3^\circ$ & $-4.6^\circ$ & $13.2^\circ$ & $11.5^\circ$ & $12.6^\circ$ \\
\hline
\end{tabular}%
}
\vspace*{-3mm}
\end{table}

For a quantitative baseline comparison, we evaluated \acrshort{lineride} against WASABI~\cite{li2023learning}, a prior method for learning agile behaviors from partial demonstrations.

The key difference is that WASABI requires a time-parameterized target trajectory, whereas \acrshort{lineride} uses only a geometric guideline, with the timing emerging naturally during training. To enable the comparison, we converted each guideline into equally spaced waypoints and treated them as a time-indexed reference for WASABI, which effectively imposes an average-speed traversal. We tested multiple waypoint intervals, corresponding to different average speeds, and trained a separate WASABI policy for each setting.

Figure~\ref{fig:baseline_wasabi} shows results on \texttt{MiniHop} and \texttt{ThreePointsTurn}. In both tasks, \acrshort{lineride} consistently outperforms WASABI. We believe this is because agile stunt behaviors require substantial speed variation within a single execution, which is difficult to capture with a fixed time-parameterized reference. Moreover, such timing is difficult for a human to specify accurately before training. These results highlight an important advantage of \acrshort{lineride}: it does not require timing information to be specified in advance.

\subsection{\acrshort{lineride} Can Be Applied to Quadrupeds}

To demonstrate the generality of our method, we applied \acrshort{lineride} to a quadruped platform and trained five diverse stunt skills: \texttt{MultiHop}, \texttt{TurnLeft}, \texttt{Circle}, \texttt{WallJump}, and \texttt{Backflip}. The resulting motions are included in the supplementary video. As shown in Fig.~\ref{fig:quadruped}, the quadruped successfully acquires all five skills using the same guideline-based task specification, suggesting that the framework is not limited to bicycle robots.

\begin{figure}\centering
\includegraphics[width=1.0\columnwidth]{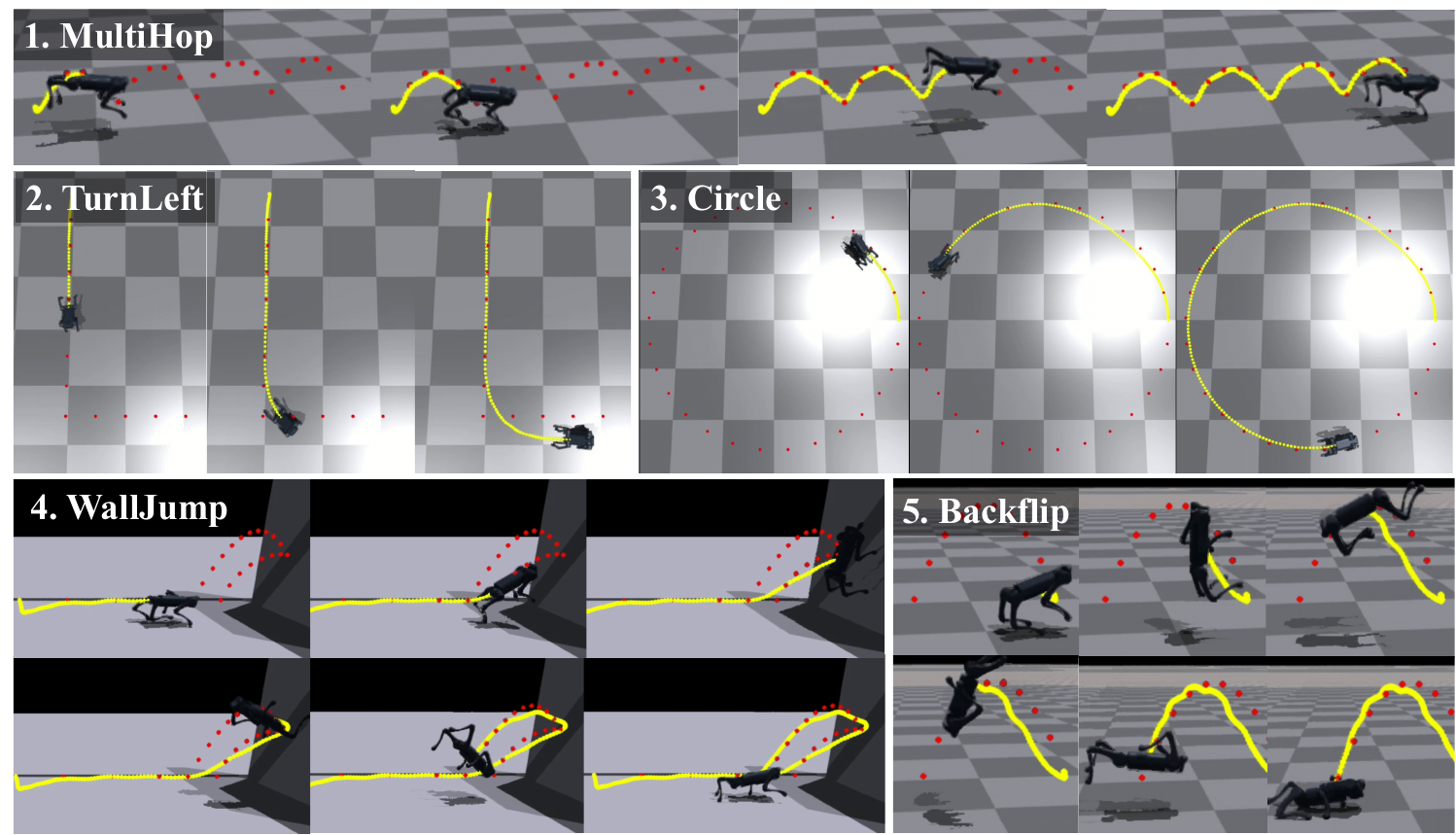}
\caption{We applied \acrshort{lineride} on quadruped for 5 stunt skills. The red dots indicate the guideline, and the yellow lines indicate the actual trajectory of the robot's base.}
\label{fig:quadruped}
\vspace*{-3mm}
\end{figure}

\section{Conclusion}

We introduced \acrshort{lineride}, a line-guided RL framework that transforms a simple user-drawn guideline and sparse key-orientations into commandable stunt behaviors on a bike-like robot. Experiments on the \gls{umv} demonstrate robust execution of five distinct skills and stable post-stunt driving, showing that lines and key-orientations provide a compact yet expressive interface that effectively connects human intent to low-level control.

\textbf{Limitations \& Future works.} Our method assumes that the user-provided guideline is at least approximately consistent with a physically realizable base trajectory, which may make it difficult to specify highly complex or long-horizon behaviors. The current stunt-mode implementation also depends on a motion-capture system for state estimation, limiting deployment to instrumented indoor environments. A promising future direction is to distill the learned policy into a phase-conditioned policy using DAGGER~\cite{ross2011reduction}, replacing explicit position input with a phase variable to enable deployment without motion capture.

\bibliographystyle{IEEEtran} \bibliography{./includes/bibliography.bib}
\end{document}